\documentclass{article}
\usepackage[dvipsnames,table]{xcolor}

\usepackage{microtype}
\usepackage{graphicx}
\usepackage{subfigure}
\usepackage{booktabs} 
\usepackage{hyperref}
\usepackage{ifthen}
\usepackage{enumitem}
\usepackage{amsmath}
\usepackage{amssymb}
\usepackage{mathtools}
\usepackage{amsthm}
\usepackage{xspace}
\usepackage{multirow}
\usepackage{colortbl}
\usepackage{tabularray}
\usepackage[table]{xcolor}
\usepackage{colortbl}
\usepackage{array}
\usepackage{makecell}
\usepackage{nicematrix}
\usepackage{xcolor}
\usepackage[textsize=tiny]{todonotes}
\usepackage[capitalize,noabbrev]{cleveref}
\usepackage{xcolor}

\theoremstyle{plain}

\theoremstyle{definition}

\theoremstyle{remark}

\usepackage[accepted]{icml2024}

\definecolor{dark2green}{RGB}{116,196,118}
\definecolor{dark2purple}{RGB}{128,125,186}
\newcommand{\first}[1]{\textbf{\textcolor{dark2green}{#1}}}
\newcommand{\third}[1]{\textbf{\textcolor{dark2purple}{#1}}}

\newcommand{\method}{\texttt{MiniMol}\xspace}

\icmltitlerunning{MiniMol: A Parameter Efficient Foundation Model for Molecular Learning}

\begin{document}

\twocolumn[
\icmltitle{MiniMol: A Parameter-Efficient Foundation Model for Molecular Learning}

\icmlsetsymbol{equal}{*}

\begin{icmlauthorlist}
\icmlauthor{Kerstin Kl\"aser}{equal,graphcore}
\icmlauthor{Błażej Banaszewski}{equal,graphcore}
\icmlauthor{Samuel Maddrell-Mander}{gc_old}
\icmlauthor{Callum McLean}{graphcore}\\
\icmlauthor{Luis M\"uller}{aachen}
\icmlauthor{Ali Parviz}{njit,mila}
\icmlauthor{Shenyang Huang}{mila}
\icmlauthor{Andrew Fitzgibbon}{graphcore}
\end{icmlauthorlist}

\icmlaffiliation{graphcore}{Graphcore}
\icmlaffiliation{gc_old}{Work carried out while at Graphcore}
\icmlaffiliation{aachen}{RWTH Aachen University}
\icmlaffiliation{njit}{New Jersey Institute of Technology}
\icmlaffiliation{mila}{Mila - Quebec AI Institute}

\icmlcorrespondingauthor{Kerstin Kl\"aser}{kerstink@graphcore.ai}
\icmlcorrespondingauthor{Błażej Banaszewski}{blazejb@graphcore.ai}

\icmlkeywords{Molecular Learning, Foundational Model, Transfer Learning}

\vskip 0.3in
]

\printAffiliationsAndNotice{\icmlEqualContribution} 

\begin{abstract}
In biological tasks, data is rarely plentiful as it is generated from hard-to-gather measurements. Therefore, pre-training foundation models on large quantities of available data and then transfer to low-data downstream tasks is a promising direction. However, how to design effective foundation models for molecular learning remains an open question, with existing approaches typically focusing on models with large parameter capacities. In this work, we propose \method, a foundational model for molecular learning with 10 million parameters. \method is pre-trained on a mix of roughly 3300 sparsely defined graph- and node-level tasks of both quantum and biological nature. The pre-training dataset includes approximately 6 million molecules and 500 million labels. To demonstrate the generalizability of \method across tasks, we evaluate it on downstream tasks from the Therapeutic Data Commons (TDC) ADMET group showing significant improvements over the prior state-of-the-art foundation model across 17 tasks. \method will be a public and open-sourced model for future research.
\end{abstract}

\section{Introduction}

\begin{figure}
    \centering
    \includegraphics[width=\columnwidth]{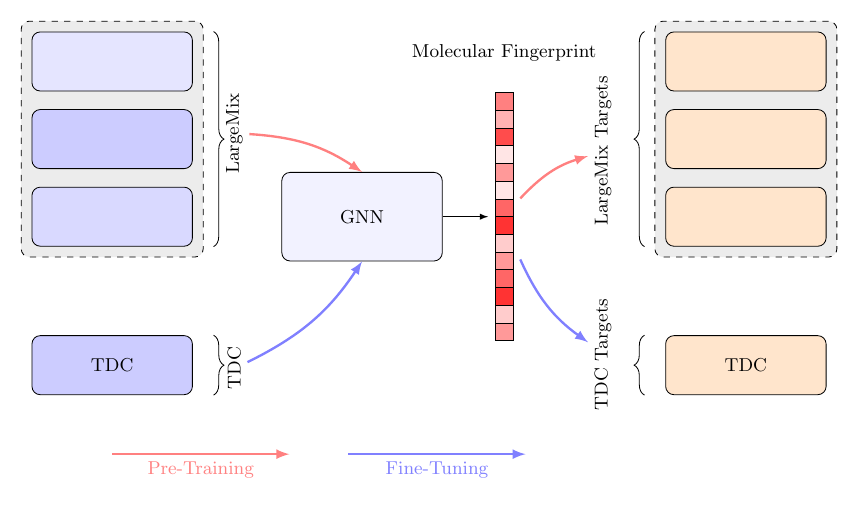}
    \caption{Workflow overview of the \method pre-training and downstream task evaluation. \method is pre-trained on the LargeMix datasets. Then \method embeddings are used as molecular fingerprints for downstasks such as those in TDC.}
    \label{fig:overview}
\end{figure}

Accurate prediction of molecular properties plays an essential role in many applications, including novel drug discovery~\cite{stokes_deep_2020, jin_deep_2021, wallach_atomnet_2015}, efficient catalyst development~\cite{zitnick_introduction_2020}, and materials design~\cite{reiser_graph_2022}.
Traditionally, Density Functional Theory~(DFT) methods~\cite{nakata2017pubchemqc} accurately compute molecular properties by physics simulation, but are computationally demanding even for small molecules, and becomes intractable in large scale of biological systems~\cite{10.3389/fchem.2023.1106495}.
Consequently, deep learning methods such as Graph Neural Networks (GNNs)~\cite{masters2023gps++,gasteiger2019directional} and graph transformers~\cite{rampavsek2022recipe} have achieved significant success in molecular representation learning.
This is demonstrated in the recent Open Graph Benchmark~(OGB) Large Scale Challenge where on the PCQM4Mv2 dataset, deep learning models produce accurate approximations of DFT while being significantly faster~\cite{lu2023highly, masters2023gps++}. In addition, by training on DFT calculations and biological tasks, ML models can predict complex biochemical properties not possible with DFT alone. 
Therefore, building foundation models that capture chemical and biological knowledge from large amounts of pre-training data and adapt it for a wide range of downstream low-data tasks is a promising next step. 

Prior work on foundational models for molecular learning has typically adopted the common practices used in computer vision and natural language processing, aiming for large model capacity and pre-training dataset size. Molecular foundation models such as MolE~\citep{mendez-lucio_mole_2022}, ChemBERTa-2 \citep{ahmad_chemberta-2_2022} and Galactica \citep{taylor_galactica_2022} directly receive SMILES strings~\cite{weininger1989smiles} of molecules as input. However, a number of equally valid SMILES strings can denote the same molecule thus SMILES string based models are unable to represent the symmetries underlying the molecular graphs.
Therefore, large amounts of pre-training data and a large model capacity is required to properly learn these symmetries. Recently, ULTRA~\citep{galkin_2023_ultra}, a foundational model for knowledge graphs, demonstrated that properly respecting the symmetries of the underlying data alleviates the need for large models in order to beat task-specific baselines. Specifically, ULTRA improves the state-of-the-art on a wide variety of knowledge graph reasoning tasks with only 177K parameters. Therefore, we leverage the permutation invariant property of GNNs to build a generally-capable and parameter-efficient foundation model for molecular fingerprinting. 

In this work, we propose \method, a parameter-efficient foundation model for molecular learning based on GNN backbone. \method is pre-trained on the Graphium \texttt{LargeMix} dataset~\cite{beaini2023towards} with around 6 million molecules and 526 million data labels. The pre-training strategy of Graphium is multi-level and multi-task meaning that over 3300 sparsely defined tasks on both graph and node level are trained jointly. \method demonstrates strong downstream task performance on the Therapeutic Data Commons~(TDC) ADMET group~\cite{huang2021therapeutics}. Figure \ref{fig:overview} shows a simplified diagram of this idea.
Our main contributions are as follows:
\begin{itemize}[leftmargin=*]
\item In this work, we propose \method, a parameter-efficient foundation model with a GINE backbone of 10~million parameters, pre-trained on around 3300 biological and quantum tasks on the graph and node level of molecules. 

\item We demonstrate that the molecular fingerprint from \method is highly transferable to downstream tasks. On the TDC benchmark, the current state of the art for a single model applied to all tasks (a foundation model) is MolE, which achieves a mean rank of 5.4, when compared against the specialized per-task models on the leaderboard. \method achieves a mean rank of 3.6 and outperforms MolE on 17 tasks.

\item We conduct a thorough performance correlation analysis between the pre-training datasets and downstream tasks. We found that \texttt{PCQM4M\_G25} dataset often has a negative correlation with downstream tasks thus highlighting the importance of understanding the correlation between pre-training tasks and downstream tasks. 

\item \method will be a public and open-sourced model for future research.  With only 10\% of the parameters of prior state-of-the-art, \method offers strong downstream performance and lower compute requirement to adapt. 
\end{itemize}

\textbf{Reproducibility:} We include the code and weight checkpoint for \method needed to reproduce the experiments as supplementary materials. 
\section{Related Work}

\subsection{Molecular Fingerprints} Traditional molecular fingerprints such as Extended Connectivity Fingerprint~(ECFP)~\cite{rogers2010extended}, RDkit fingerprints~\cite{landrum2013rdkit} and MAP4~\cite{capecchi2020one} are designed for molecular characterization, similarity searching, and structure-activity modelling, with wide applications in drug discovery. However, they encode the presence of particular substructures within the molecule and should be manually customized for specific applications. In addition, it is shown that different types of fingerprints perform better for specific categories of molecules. For example, substructure fingerprint~\cite{kim2021exploring} has the best performance on small molecules such as drugs while atom-pair fingerprints~\cite{awale2014atom} are best suited for large molecules such as peptides. Even SOTA fingerprints can suffer from embedding collisions due to how sub-structures are resolved \cite{probst_reaction_2022}. From large molecular datasets, foundation models aim to learn a universal and descriptive molecular representation as practical molecular fingerprints for downstream tasks. Our \method model generates strong fingerprints for various tasks on the TDC benchmark as indicated by downstream task performance. 

\subsection{Foundational Models in ML.} In Natural Language Processing~(NLP) and Computer Vision~(CV), foundation models have achieved significant progress, especially for Large Language Models~(LLMs)~\cite{achiam2023gpt}. Foundation models are often pre-trained on a mixture of data across a wide range of tasks. Downstream tasks with low amounts of data can then be solved using low-resource, inexpensive fine-tuning~\cite{tian2023fine,borzunov2023distributed}. 
Multi-modal datasets have been used with LLMs to align a single model representation across domains \cite{team2023gemini,betker2023improving}. 
In most areas, these general properties have typically been emergent from extremely large models.

\subsection{Foundational Models in Molecular Learning.} Initial unsupervised transformer-based designs most closely replicate work in NLP~\cite{honda_smiles_2019, wang_smiles-bert_2019,honda_smiles_2019}, representing molecules as SMILES strings~\cite{weininger_smiles_1988}. These models leverage extremely large but low-fidelity unlabelled molecule datasets. Early FMs can achieve strong results on some small, out-of-domain tasks. However, generalizability to a wide variety of tasks remains limited~\cite{zhu_dual-view_2021, liu_towards_2023, mendez-lucio_mole_2022,luo_molfm_2023}. Recent state-of-the-art in molecular property prediction employs geometric deep learning, often with an MPNN or Graph Transformer, trained with supervised labels of molecular properties~\cite{ying_transformers_2021, velickovic_graph_2017, dwivedi_generalization_2020}. Recent work by ~\cite {masters_gps_2023} has shown promising potential for MPNN scaling in-depth and total parameters for predicting the quantum properties of molecules. Recent work COATI~\cite{kaufman_coati_2023} is based on SMILES and point clouds, utilizing a multi-modal encoder-decoder scheme designed for molecule regression tasks. Significantly, \citet{shoghi_molecules_2023} demonstrates the importance of multi-task pre-training on a range of length scales for low-resource fine-tuning and extracting learned representations for molecular property prediction (MPP). In this work, we use graph-based representations of molecules, which provide rich chemical and structural information. 
\method is pre-trained on a large number of molecular properties of both quantum and biological tasks while demonstrating strong performance across many downstream tasks. 
\begin{figure*}[t!]
    \centering
    \includegraphics[width=0.9\textwidth]{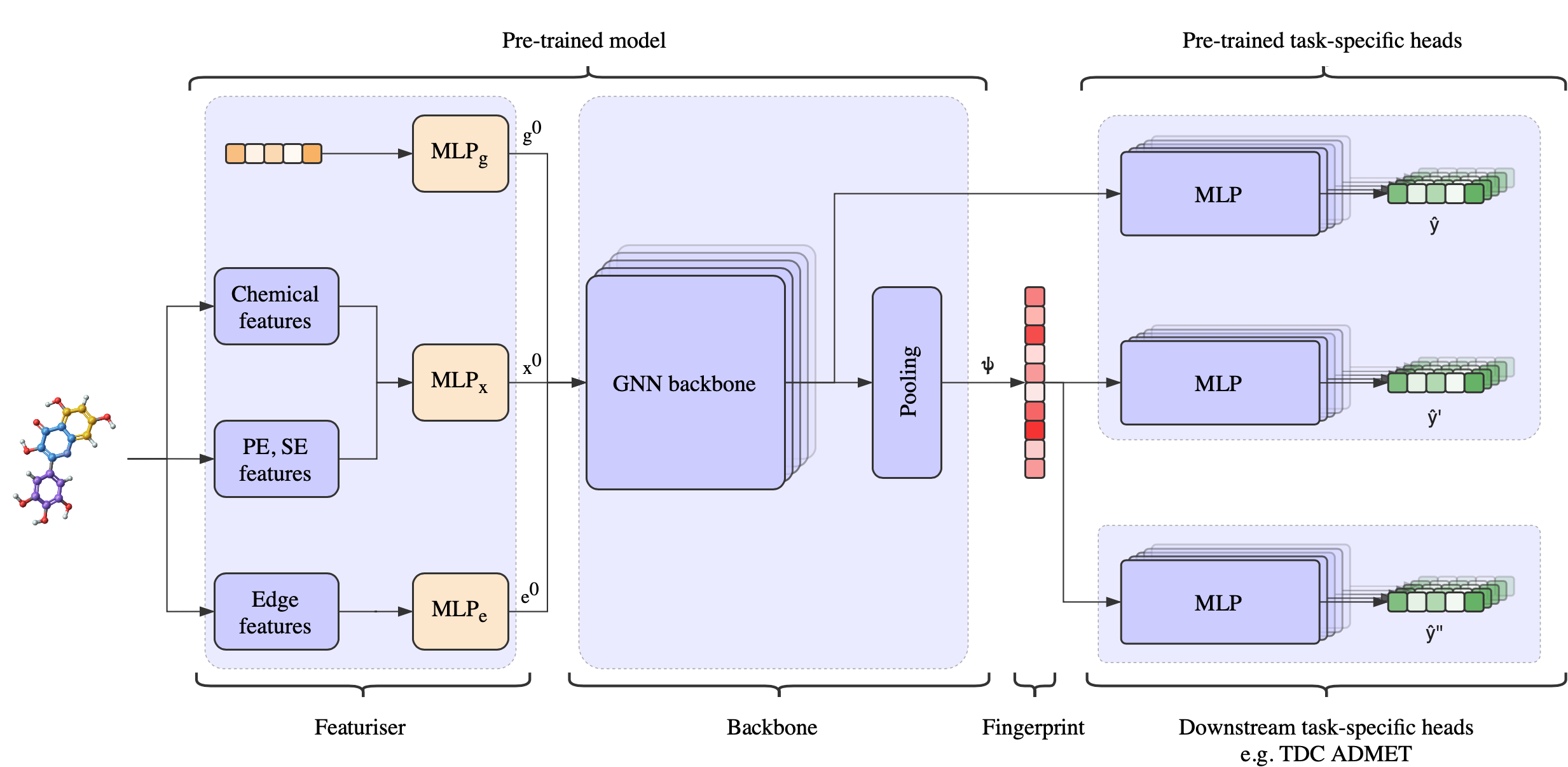}\\[-1ex]
    \caption{Schematic of the architecture of \method. An example molecule is featurized in the first block. Node feature vectors are created by combining chemical features with positional and structural encodings, edge features are generated using RDKit and a random initial global vector is generated. Each initial vector is processed with a separate embedding MLP. The backbone of the model is a stack of MPNN layers, which output the molecular fingerprint $\psi$ after pooling. The pre-pooling output is used for pre-training on node-level tasks, in our case, PCQM4M\_N4. The fingerprint $\psi$ is used either for pre-training multiple graph-level task heads or as an input to the downstream tasks, including the full set of \texttt{ADMET} tasks from the TDC benchmarks. }
    \label{fig:pipeline}
\end{figure*}

\section{Method}

Here, we present our architecture for pre-training on the \texttt{LargeMix} datasets~\citep{beaini_towards_2023}, extracting fingerprints and subsequently fine-tuning to downstream tasks (see Fig. \ref{fig:pipeline}).

\subsection{Molecular Representation}
Each molecule is modelled as a graph $\mathcal{G}$ with $N$ nodes representing the atoms and $M$ edges representing the bonds. We denote the set of edges with $\mathcal{E}$.
The atom and bond features are generated using RDKit, providing a set of categorical and floating values, and the atomic features are concatenated with positional and structural embeddings.
From \cite{masters_gps_2023, rampasek_recipe_2022} the Laplacian eigenvectors and eigenvalues, and the random walk probabilities, were found to be most beneficial.
The input node feature vectors are the concatenation of these features 
\begin{equation*}
    X^0 = \left[
    X^{\mathrm{atom}} |
    X^{\mathrm{LapVec}} | 
    X^{\mathrm{LapVal}} |
    X^{\mathrm{RW}} 
    \right],
\end{equation*}
and edge features are the bond features $E^0 
 = \left[E^{\mathrm{bond}}\right]$. 
 A global node is added to each graph, providing an additional connection to every node. It was shown in \cite{li_learning_2017}) that the global node dramatically improves graph-level representation. 
 This acts both as routing between otherwise distant portions of the graph and as a readout node for the graph property. It is initialized with a random vector. 
 Each of the nodes, edges, and global features are initially embedded into the model dimensions using a two-layer MLP each (eq \ref{eq:x} - \ref{eq:g}). 

\begin{equation}
    x^{0} = \texttt{MLP}_x\left(X^0\right) \quad \in \mathbb{R}^{N\times d_{\mathrm{node}}}
    \label{eq:x}
\end{equation}
\begin{equation}
    e^{0} = \texttt{MLP}_e\left(E^0\right) \quad \in \mathbb{R}^{M\times d_{\mathrm{edge}}}
    \label{eq:e}
\end{equation}
\begin{equation}
    g^{0} = \texttt{MLP}_g(\texttt{rand}_g\left(0\right)) \quad \in \mathbb{R}^{d_{\mathrm{global}}}
\label{eq:g}
\end{equation}

\subsection{Model}\label{sec:method_model}
Given the initial node, edge and graph embeddings we update them through multiple layers of message-passing to obtain final node embeddings
\begin{equation*}
    x^\text{final} = \texttt{GNN}(x^0, e^0, g^0),
\end{equation*}
where $\texttt{GNN}$ is a chosen GNN backbone. As described later in \Cref{sec:experiments}, we try three different backbone GNNs, namely GCN \citep{kipf2016semi}, GINE \citep{hu_strategies_2020,xu_how_2019} and MPNN++ \citep{masters2023gps++}. We briefly describe the different architectures here.

\paragraph{GCN} The GCN \citep{kipf_semi-supervised_2016} incorporates only the node embeddings. Concretely, the $\ell$-th layer of a GCN is defined as
\begin{equation*}
    x_i^{\ell +1} = \sum_{j \in \mathcal{N}(i) \cup \{i\}} \dfrac{1}{\sqrt{d_i d_j}} x_j^{\ell},
\end{equation*}
where $\mathcal{N}(i)$ denotes the set of neighbors of $i$ and $d_i$ and $d_j$ denote the degree of nodes $i$ and $j$, respectively.

\paragraph{GINE} GINE is an extension of GIN \citep{xu_how_2019} to additionally incorporate edge embeddings \citep{hu_strategies_2020}. Concretely, the $\ell$-th layer of GINE is defined as
\begin{equation*}
   x_i^{\ell +1} = \texttt{MLP} \Big( (1 - \epsilon) \cdot x_i^{\ell} \sum_{j \in \mathcal{N}(i)} \texttt{ReLU}(x_j^{\ell} + e^\ell_{ij}) \Big), 
\end{equation*}
where $\epsilon$ is a learnable scalar. Compared to GCN, GINE is more expressive as it has the same expressive power for graph isomorphism test as the $1$-Weisfeiler-Leman algorithm \citep{xu_how_2019}. Further, GINE has been used previously for pre-training graph neural networks \citep{hu_strategies_2020}.

\paragraph{MPNN++} The Message Passing Neural Network~(MPNN) module in~\cite{masters2023gps++} called MPNN++, is designed to incorporate and update node-, edge- and graph-level embeddings. The MPNN++ is a complex architecture as it aggregates messages from adjacent nodes, incident edges and global graph-level embeddings. In addition, the MPNN++ uses layer-wise skip connections for node-, edge- and graph-level embeddings; see \Cref{sec:mpnn_details} for a detailed description of the MPNN++ layer. We include MPNN++ as a backbone in our experiments. 

\subsection{Pre-training}

\method is jointly pre-trained with many supervised tasks on both the graph and node levels. 
The total loss minimized during training is a summation of each of the pre-training tasks, accounting for label sparsity per molecule.
There is a rich literature on how to combine losses for multi-task learning~\cite{crawshaw2020multi}, although in this case, the tasks are a combination of regression and classification, meaning the losses are not naturally commensurate.  The mean absolute error (MAE) loss is used for the PCQM dataset (N4 and G25 tasks), the binary cross-entropy (BCE) loss is used for the PCBA tasks, and the hybrid cross-entropy (HCE) loss from \cite{beaini_towards_2023} is used for the L1000 datasets. Concretely, we compute the final loss $\mathcal{L}$ as
\begin{equation*}
   \mathcal{L} = \mathcal{L}_\text{L1000} + \mathcal{L}_\text{PCBA} + \mathcal{L}_\text{N4} + \frac{1}{k} \mathcal{L}_\text{G25}.
\end{equation*}
where $k$ is a scaling constant and we set $k=5$ to account for data imbalance on the G25 dataset (see more discussion in \Cref{sec:experiments}). 

\subsection{Downstream Fingerprinting} 
For downstream tasks, we generate the global embeddings of the final layer of \method from a given molecule which is also referred to as molecular fingerprints. This is more compute efficient and easy to use when compared to fine-tuning the entire model from end-to-end. In addition, generating meaningful molecular fingerprints for downstream tasks is an important aspect of the foundation model of molecular learning.

More specifically, we first extract fingerprints for all unique molecules in a given downstream task and subsequently use the fingerprints as molecular representation to train a small Multilayer Perceptron~(MLP) for making task-specific predictions.
To generate the fingerprints from \method, we compute the final node-, edge- and graph-level embeddings as described in \Cref{sec:method_model} and subsequently obtain fingerprints by pooling the final node embeddings, e.g., via max pooling, obtaining
\begin{equation*}
    \psi = \sum_{i \in \mathcal{G}} x^\text{final}_i
\end{equation*}
for the fingerprint vector $\psi$.

There are several advantages of using molecular fingerprints for downstream tasks when compared to end-to-end fine-tuning. First, the above procedure allows a significantly more efficient way to train a model on low-data downstream tasks. Molecular fingerprints are pre-computed for a given set of molecules with a single forward pass from the model and then used for many downstream tasks or hyperparameter sweeps. Second, the downstream model is less likely to overfit when only the downstream MLP is trainable as there are fewer trainable parameters. Next, the use of molecular fingerprints for downstream tasks in this way matches existing workflows in the bio-chemistry domain increasing the practical utility of the method. Finally, as fingerprints are simply embedding vectors, practitioners are not required to access the architecture or weight of the pre-trained model thus only the expertise to train a task head (or MLP) is required.

\section{Experimental Details}\label{sec:experiments}
In our experiments, we pre-train \method on \texttt{LargeMix} \citep{beaini_towards_2023} for various GNN backbones and subsequently fine-tune to all 22~tasks in the ADMET Group of the TDC benchmark.

\subsection{Pre-training}\label{sec:exp_pre_training} 
\method uses the \texttt{LargeMix} datasets from~\cite{beaini_towards_2023} for pre-training, consisting of approximately 6M molecules and a total of 526M targets, which have been summarized in Table ~\ref{tab:datasets}. The datasets are described below.

\textbf{\texttt{PCQM4M\_G25\_N4}.} This dataset contains 3.8M molecules from the PCQM dataset~\cite{hu2021ogb}, from the OGB-LSC challenge. The dataset consists of quantum chemistry calculations for 25~molecular graph-level properties, and 4 node-level properties per atom, resulting in about 400M labelled data points.

\textbf{\texttt{PCBA}.} This dataset contains 1.5M molecules from the OGBG-PCBA dataset~\cite{hu2020open}. This bioassay dataset, derived experimentally from high-throughput screening methods, details the impact of the molecules on living cells across 1328 sparse labels. This results in about 100M labelled data points.

\textbf{\texttt{L1000\_VCAP} and \texttt{L1000\_MCF7}.} These datasets contain 26k molecules from the L1000 dataset~\cite{subramanian2017next} which details the change to gene expression profiles and cellular processes when exposed to the molecules in the dataset across about 1000~labels and 26M data points. 

\begin{table}[h!]
\centering
\caption{Overview of the datasets in LargeMix. \vspace{5pt}}
\resizebox{0.48\textwidth}{!}{ 
    \begin{tabular}{lcccc}
    \hline
    \textbf{Dataset} & \textbf{\# Molecules} & \textbf{\# Labels} & \textbf{\# Data Points} & \textbf{\% of All Data Points} \\
    \hline
    PCQM4M\_G25 & 3.81M & 25 (G) & 93M & 17\% \\
    PCQM4M\_N4 & 3.81M & 4 (N) & 197.7M & 37\% \\
    PCBA\_132B & 1.56M & 1328 (G) & 224.4M & 41\% \\
    L1000\_VCAP & 15K & 978 (G) & 15M & 3\% \\
    L1000\_MCF7 & 12K & 978 (G) & 11M & 2\% \\
    \hline
    \end{tabular}
}
\label{tab:datasets}
\end{table}

These diverse labels from fundamental quantum chemistry properties to macro-scale cellular impact encourage a single general representation of the molecule suitable for downstream tasks. The combined \texttt{LargeMix} contains multiple task labels per molecule. The datasets only partially overlap thus requiring the model to generalize across domains from sparse labels on molecules. Following~\cite{mendez-lucio_mole_2022}, we filter out molecules with more than 100~heavy atoms. In addition, we remove molecules in the ADMET group test sets from our pre-training data to avoid potential leakage of test labels (7\% of MCF7, 4\% of VCAP, 0.6\% of PCBA, 0.07\% of PCQM4M\_G25\slash N4). During pre-training we split the dataset into 92\% training, 4\% validation, and 4\% test data.

To cover a range of GNN backbones with increasing complexity, we pre-train GCN, GINE, and finally MPNN++ models and subsequently evaluate their downstream performance on the TDC ADMET group datasets. Each model consists of 16~GNN layers with hidden dimensions adjusted such that all models have around 10M parameters. We train each model for 100~epochs using the Adam optimizer, with a maximum learning rate of $3e^{-4}$, 5~warm-up epochs and linear learning rate decay.

\subsection{Benchmarking on TDC ADMET Group} \label{sec:exp_fingerprints}
Here, we describe the ADMET group of the TDC benchmark which we use to evaluate the downstream performance of \method.
The Therapeutics Data Commons (TDC)~\cite{huang2021therapeutics} is a platform designed to facilitate the assessment and development of AI methods in drug discovery. It particularly emphasizes identifying the most effective AI techniques for this purpose.
Within TDC, the ADMET Benchmark Group specializes in single-instance prediction, offering a standardized suite of 22 datasets for molecular property prediction. These datasets vary in size, ranging from 475 to 13,130 molecules, and encompass tasks in both regression and classification. The datasets span a breadth of molecular properties, categorized into Absorption, Distribution, Metabolism, Excretion, and Toxicity. To ensure fair comparability, scaffold splits are employed, with 80\% of data for training and 20\% for testing.

A diverse array of models, including random forests, GNNs, and CNNs, are evaluated on these tasks. Their performance is showcased on the \href{https://tdcommons.ai/benchmark/admet_group/overview/}{TDC leaderboard} which includes various models trained with SMILES or other encoding strategies. To benchmark \method, we first build an ensemble by training a distinct model on each fold in 5-fold cross-validation.
While building the ensemble, the best epoch is selected based on validation loss, and to distinguish which ensemble to select for testing (e.g. while choosing one out of the sweep), the ensemble's mean validation metric is used. Final test scores are derived from the top ensemble, with error bars reported from five trials (see Appendix~\ref{sec:ensembling_strategy} for pseudo code). Table~\ref{tab:largemix_table} presents the performance of three GNN architectures (GCN, GINE, MPNN++) across various datasets and two computational budgets for sweeping hyperparameters of the dataset-specific models. Since our fingerprinting approach permits fast evaluation of downstream predictors, we conduct extensive hyperparameter sweeping across all tasks (see Appendix \ref{sec:hyperparameter_selection} for more details about the hyperparameter selection). For CPU-only runs, training a downstream model only takes 1 to 10 minutes per model per dataset.

\begin{table*}[t!]
\centering
\caption{Results on downstream evaluation of \method (GINE) with max pooling on TDC ADMET benchmarks, and comparison to the TDC leaderboard and MolE. The rank is determined for each dataset individually, on a set of 7 scores, which include the test results from the TOP5 leaderboard, MolE and \method. The best result is shown in \first{green} and the top three results are highlighted in \third{purple}.\vspace{10px}}
\label{tab:admet_original}
\resizebox{\textwidth}{!}{
\begin{NiceTabular}{|l|l|r|r|l|l|c|l|c|}
\toprule
& \multicolumn{3}{c}{TDC Dataset} & \makecell{Leaderboard \\ Jan. 2024}
& \multicolumn{2}{c}{MolE} & \multicolumn{2}{c}{\method~(GINE)}\\
\midrule
    & Name             & Size             & Metric    & SOTA Result                    & Result                & Rank & Result               & Rank \\
\midrule                 
\multirow{6}{*}{\sc \rotatebox{90}{ \scriptsize Absorption}}
    & Caco2 Wang            & 906       & MAE~($\downarrow$) & \third{0.276 \tiny{$\pm$ }.005} & 0.310 \tiny{$\pm$ }.010 & 6            & 0.324 \tiny{$\pm$ }.012           & 7 \\
    & Bioavailability Ma    & 640       & AUROC~($\uparrow$)    & \first{0.748 \tiny{$\pm$ }.033} & 0.654 \tiny{$\pm$ }.028 & 7         & 0.699 \tiny{$\pm$ }.008           & 6 \\
    & Lipophilicity AZ      & 4,200      & MAE~($\downarrow$) & \third{0.467 \tiny{$\pm$ }.006} & \third{0.469 \tiny{$\pm$ }.009} & 3   & \first{0.455 \tiny{$\pm$ }.001}   & 1 \\
    & Solubility AqSolDB    & 9,982      & MAE~($\downarrow$) & \third{0.761 \tiny{$\pm$ }.025} & 0.792 \tiny{$\pm$ }.005 & 5           & \first{0.750 \tiny{$\pm$ }.012}   & 1 \\
    & HIA Hou               & 578       & AUROC~($\uparrow$)  & \third{0.989 \tiny{$\pm$ }.001} & 0.963 \tiny{$\pm$ }.019 & 7           & \first{0.994 \tiny{$\pm$ }.003}   & 1 \\ 
    & Pgp Broccatelli       & 1,212      & AUROC~($\uparrow$) & \third{0.938 \tiny{$\pm$ }.002} & 0.915 \tiny{$\pm$ }.005 & 7           & \first{0.994 \tiny{$\pm$ }.002}   & 1 \\
\midrule
\multirow{3}{*}{\sc \rotatebox{90}{\scriptsize Distrib.}}
    & BBB Martins           & 1,975      & AUROC~($\uparrow$) & \third{0.920 \tiny{$\pm$ }.006} & 0.903 \tiny{$\pm$ }.005 & 7               & \first{0.923 \tiny{$\pm$ }.002}   & 1 \\
    & PPBR AZ               & 1,797      & MAE~($\downarrow$) & \third{7.526 \tiny{$\pm$ }.106} & 8.073 \tiny{$\pm$ }.335 & 6               & 7.807 \tiny{$\pm$ }.188           & 4 \\
    & VDss Lombardo         & 1,130      & Spearman~($\uparrow$) & \first{0.713 \tiny{$\pm$ }.007} & \third{0.654 \tiny{$\pm$ }.031} & 3    & 0.570 \tiny{$\pm$ }.015           & 7 \\
\midrule
\multirow{6}{*}{\sc \rotatebox{90}{\scriptsize Metabolism}}
    & CYP2C9 Veith      & 12,092 & AUPRC~($\uparrow$) & \first{0.859 \tiny{$\pm$ }.001} & 0.801 \tiny{$\pm$ }.003 & 5           & 0.819 \tiny{$\pm$ }.001           & 4 \\
    & CYP2D6 Veith      & 13,130 & AUPRC~($\uparrow$) & \first{0.790 \tiny{$\pm$ }.001} & 0.682 \tiny{$\pm$ }.008 & 6           & 0.718 \tiny{$\pm$ }.003           & 5 \\
    & CYP3A4 Veith      & 12,328 & AUPRC~($\uparrow$) & \first{0.916 \tiny{$\pm$ }.000} & 0.867 \tiny{$\pm$ }.003 & 7           & 0.878 \tiny{$\pm$ }.001           & 5 \\
    & CYP2C9 Substrate  & 666   & AUPRC~($\uparrow$) & \third{0.441 \tiny{$\pm$ }.033} & \third{0.446 \tiny{$\pm$ }.062} & 2    & \first{0.481 \tiny{$\pm$ }.013}   & 1 \\
    & CYP2D6 Substrate  & 664   & AUPRC~($\uparrow$) & \first{0.736 \tiny{$\pm$ }.024} & 0.699 \tiny{$\pm$ }.018 & 7            & \third{0.726 \tiny{$\pm$ }.006}   & 2 \\
    & CYP3A4 Substrate  & 667   & AUROC~($\uparrow$) & \third{0.667 \tiny{$\pm$ }.019} & \first{0.670 \tiny{$\pm$ }.018} & 1    & 0.644 \tiny{$\pm$ }.006           & 6  \\
\midrule
\multirow{3}{*}{\sc \rotatebox{90}{\scriptsize Excret.}}
    & Half Life Obach       & 667   & Spearman~($\uparrow$)  & \first{0.576 \tiny{$\pm$ }.025} & 0.549 \tiny{$\pm$ }.024 & 4    & 0.493 \tiny{$\pm$ }.002           & 7 \\
    & Clearance Hepatocyte  & 1,102  & Spearman~($\uparrow$)  & \first{0.536 \tiny{$\pm$ }.020} & 0.381 \tiny{$\pm$ }.038 & 7   & 0.448 \tiny{$\pm$ }.006           & 4 \\
    & Clearance Microsome   & 1,020  & Spearman~($\uparrow$)  & \third{0.630 \tiny{$\pm$ }.010} & 0.607 \tiny{$\pm$ }.027 & 6   & \first{0.652 \tiny{$\pm$ }.007}   & 1 \\
\midrule
\multirow{4}{*}{\sc \rotatebox{90}{\scriptsize Toxicity}}
    & LD50 Zhu              & 7,385  & MAE~($\downarrow$)    & \first{0.552 \tiny{$\pm$ }.009} & 0.823 \tiny{$\pm$ }.019 & 7        & \third{0.588 \tiny{$\pm$ }.010}   & 3 \\
    & hERG                  & 648   & AUROC~($\uparrow$)  & \first{0.880 \tiny{$\pm$ }.002} & 0.813 \tiny{$\pm$ }.009 & 7           & 0.849 \tiny{$\pm$ }.007           & 6 \\
    & Ames                  & 7,255  & AUROC~($\uparrow$)  & \third{0.871 \tiny{$\pm$ }.002} & \first{0.883 \tiny{$\pm$ }.005} & 1  & 0.856 \tiny{$\pm$ }.001           & 5 \\
    & DILI                  & 475   & AUROC~($\uparrow$)  & \third{0.925 \tiny{$\pm$ }.005} & 0.577 \tiny{$\pm$ }.021 & 7           & \first{0.944 \tiny{$\pm$ }.007}   & 1 \\
\midrule
\multicolumn{5}{r}{ TDC Leaderboard Mean Rank:} & \multicolumn{2}{r}{5.4~~~} & \multicolumn{2}{r}{\bf 3.6~~~{}}  \\
\bottomrule
\end{NiceTabular}
}
\end{table*}

\section{Empirical Results}
Here, we present our experimental results for both pre-training and fine-tuning on fingerprints.

\subsection{Pre-training on LargeMix}
We present our pre-training results for \method with GCN, GINE and MPNN++ as backbone GNNs on \texttt{LargeMix} in \Cref{tab:largemix_table}. Here, we observe that pre-training performance is only marginally affected by the choice of the backbone GNN. Moreover, the pre-training performance of a given GNN backbone also varies with tasks. For example, on the graph- and node-level regression tasks of \texttt{PCQM4M\_G25} and \texttt{PCQM4M\_N4} the MPNN++ backbone performs consistently better while GINE performs best on the classification tasks of \texttt{PCBA\_1328}. On \texttt{L1000\_VCAP} we observe roughly comparable results across all metrics with a slight advantage of GINE in terms of AUROC. Finally, on \texttt{L1000\_MCF7}, GCN and GINE are largely on par with the MPNN++ showing worse performance in terms of AUROC. Most importantly, no single backbone is superior to the other two. Next, we present our results of fine-tuning these models to the ADMET group benchmark of TDC using our fingerprinting approach. 

\begin{table}[t]
\centering
\vspace{-0.4ex}
\caption{Results for GNN 10M baselines on \textsc{LargeMix} dataset. We report performance metrics on the test set for each dataset in \textsc{LargeMix} separately. The best scores per metric per dataset are marked in bold.}
\vspace{10pt}
\label{tab:largemix_table}

\resizebox{0.45\textwidth}{!}{ 	
\begin{NiceTabular}{l|l|cccc}
\toprule
&  \quad& \multicolumn{4}{c}{Model} \\
Dataset     & Metric     & GCN & GINE & MPNN \\
                         \midrule
\multirow{3}{*}{\sc Pcqm4m\_g25}
    & MAE $\downarrow$   & 0.218 & 0.208 & \textbf{0.200} \\
    & Pearson $\uparrow$   & 0.884 & 0.889 & \textbf{0.892} \\
    & $R^2$ $\uparrow$  & 0.790 & 0.799 & \textbf{0.803}\\
\midrule
\multirow{3}{*}{\sc Pcqm4m\_n4}
    & MAE $\downarrow$   & 0.025 & 0.022 & \textbf{0.021} \\
    & Pearson $\uparrow$  & 0.975 & 0.979 & \textbf{0.980} \\
    & $R^2$ $\uparrow$  & 0.952 & 0.959 & \textbf{0.961} \\
\midrule
\multirow{3}{*}{\sc Pcba\_1328}
    & CE $\downarrow$   & 0.033& 0.033 & 0.033 \\ 
    & AUROC $\uparrow$  & 0.777 & \textbf{0.784} & 0.782 \\ 
    & AP $\uparrow$  & 0.286 &  \textbf{0.302} & 0.287 \\
\midrule
\multirow{3}{*}{\sc L1000\_vcap}
    & CE $\downarrow$   & \textbf{0.061} & \textbf{0.061} & \textbf{0.061} \\ 
    & AUROC $\uparrow$  & 0.500 & \textbf{0.514} & 0.500  \\
    & AP $\uparrow$  & 0.504 &  0.504   &   \textbf{0.506} \\
\midrule
\multirow{3}{*}{\sc L1000\_mcf7}
    & CE $\downarrow$   & 0.059 &  \textbf{0.058} & 0.059  \\
    & AUROC $\uparrow$  & \textbf{0.533} & 0.531  & 0.519 \\
    & AP $\uparrow$  & 0.513 &   \textbf{0.516} & 0.514 \\
\bottomrule
\end{NiceTabular}
}

\end{table}

\begin{table}[t]
    \centering
    \caption{The effect of specific GNN architectures in the backbone of the fingerprinting model on the downstream performance. The rank is determined for each dataset individually, on a set of 7~scores, which include the test results from the TOP5 TDC leaderboard, MolE and \method. Here, all models used sum pooling, whereas our best model uses max pooling.}
    \label{tab:models_on_tdc}
    \vspace{10pt}

    \resizebox{0.5\textwidth}{!}{ 	
        \begin{NiceTabular}{@{}r|c|c|c@{}}
        \toprule
            \method backbone  & Mean Rank    & \# Top1 Results & \# Top3 Results \\ 
        \midrule
        MPNN++    & 4.8           & 3     & 6 \\
        GCN       & 4.4           & 4     & 8 \\
        GINE      & 3.9           & 5     & 10 \\
        \bottomrule
        \end{NiceTabular}
    }
\end{table}

\subsection{Downstream performance on TDC}
Here, we present our fine-tuning results on the ADMET group datasets of TDC; see \Cref{tab:models_on_tdc} for a comparison of different backbone GNNs in terms of downstream performance.
GINE demonstrates a significant empirical advantage as the GNN backbone for downstream tasks. Thus, in \Cref{tab:admet_original}, we compare \method (GINE), to the TDC leaderboard and the current state-of-the-art, MolE~\cite{mendez-lucio_mole_2022}. \method with GINE backbone, achieves top 1 performance on 8 tasks, setting a new state-of-the-art on these datasets. Moreover, \method (GINE) achieves top-3 performance on 10 tasks. Therefore, \method (GINE) is shown to be a versatile model across a wide range of tasks, competing with or exceeding the performance of the best task-specialized architectures. We report TDC leaderboard results up until January 2024. In addition, \method (GINE) outperforms MolE on 17 datasets, indicating that with only 10\% of the parameters, our \method approach is favourable to MolE in downstream performance across many molecular tasks. For our best model, we explored different pooling methods, see Appendix ~\ref{sec:pooling_experiments}.

In Table ~\ref{tab:more_models_on_tdc}, we compare \method to other fingerprinting methods using the same evaluation downstream adaptation method as ours, so the only difference is the quality of generated fingerprints.

\begin{table}[t]
\centering
\caption{Comparison of \method to other molecular fingerprinting models using the same evaluation method as ours, including ensembles.\vspace{10pt}}
\resizebox{0.5\textwidth}{!}{ 	
    \begin{tabular}{lcccc}
        \hline
        \textbf{Model} & \textbf{Mean rank} & \textbf{$>$MoIE} & \textbf{TOP1} & \textbf{TOP3} \\ \hline
        \method                         & 3.6 & 17 & 8 & 10 \\
        AGBT~\cite{chen2021algebraic}   & 5.7 & 10 & 1 & 4  \\
        MolFormer~\cite{ross2022large}  & 5.7 & 7  & 0 & 4  \\ 
        BET~\cite{chen2021extracting}   & 6.1 & 7  & 1 & 2  \\ \hline
    \end{tabular}
}
\label{tab:more_models_on_tdc}
\end{table}

\begin{table*}[t]
\label{tab:spearman}
\centering
\caption{Correlation analysis (Spearman's rho) between pre-training validation and downstream performance.
The green color indicates positive correlation , and red a negative correlation.
Results with a p-value over 10\% are blank.
See \S{\ref{sec:correlation}} for discussion.
\label{tab:correlation_analysis}
\vspace{8pt}} 
\resizebox{1.0\textwidth}{!}{
\def\csc#1{%
\ifdim#1pt>0.6pt\cellcolor{green!60}#1\else
\ifdim#1pt>0.4pt\cellcolor{green!40}#1\else
\ifdim#1pt>0.2pt\cellcolor{green!20}#1\else
\ifdim#1pt<-0.6pt\cellcolor{red!60}#1\else
\ifdim#1pt<-0.4pt\cellcolor{red!40}#1\else
\ifdim#1pt<-0.2pt\cellcolor{red!20}#1\else
\cellcolor{gray!20}\fi\fi\fi\fi\fi\fi
}
\begin{NiceTabular}{l|l|c|c|c|c|c|c|c|c|c|c}
\toprule
 \multirow{2}{*}{Dataset} & \multirow{2}{*}{Metric} & \multirow{2}{*}{Overall Loss} 
 & \multicolumn{2}{c}{\texttt{L1000\_MCF}} & \multicolumn{2}{c}{\texttt{L1000\_VCAP}} & \multicolumn{2}{c}{\texttt{PCBA}} & \multicolumn{2}{c}{\texttt{PCQM4M\_G25}}        & \multicolumn{1}{l}{\texttt{PCQM4M\_N4}}  \\

&  &                                                & Loss  & AUROC           & Loss  & AUROC            & Loss  & AUROC      & Loss  & MAE & Loss (MAE)                 \\ 
\midrule
Caco2 Wang                & MAE                     & \csc{0} & \csc{0} & \csc{0.590} & \csc{0} & \csc{0.651} & \csc{0.762} & \csc{0.718} & \csc{0} & \csc{0} & \csc{0}\\
Bioavailability Ma        & AUROC                   & \csc{0} & \csc{0} & \csc{0} & \csc{0.397} & \csc{0} & \csc{0} & \csc{0} & \csc{0} & \csc{0} & \csc{0}\\
Lipophilicity AZ          & MAE                     & \csc{0.459} & \csc{0.561} & \csc{0.568} & \csc{0} & \csc{0.539} & \csc{0.683} & \csc{0.627} & \csc{0} & \csc{-0.389} & \csc{0}\\
Solubility AqSolDB        & MAE                     & \csc{0.434} & \csc{0.472} & \csc{0.588} & \csc{0} & \csc{0.7} & \csc{0.739} & \csc{0.704} & \csc{0} & \csc{0} & \csc{0}\\
HIA Hou                   & AUROC                   & \csc{0.427} & \csc{0.489} & \csc{0.603} & \csc{0.382} & \csc{0.548} & \csc{0.768} & \csc{0.645} & \csc{0} & \csc{-0.337} & \csc{0}\\
Pgp Broccatelli           & AUROC                   & \csc{0.444} & \csc{0.577} & \csc{0} & \csc{0} & \csc{0.361} & \csc{0.497} & \csc{0} & \csc{0} & \csc{-0.387} & \csc{0}\\
BBB Martins               & AUROC                   & \csc{0.634} & \csc{0.436} & \csc{0.583} & \csc{0} & \csc{0.378} & \csc{0.481} & \csc{0.483} & \csc{0.364} & \csc{-0.492} & \csc{0}\\
PPBR AZ                   & MAE                     & \csc{0} & \csc{0.476} & \csc{0} & \csc{0.436} & \csc{0} & \csc{0} & \csc{0} & \csc{0} & \csc{0} & \csc{0}\\
VDss Lombardo             & Spearman                & \csc{0.408} & \csc{0} & \csc{0} & \csc{0} & \csc{0.343} & \csc{0.351} & \csc{0} & \csc{0} & \csc{0} & \csc{0}\\
CYP2C9 Veith              & AUPRC                   & \csc{0.432} & \csc{0} & \csc{0.649} & \csc{0} & \csc{0.711} & \csc{0.747} & \csc{0.829} & \csc{0.557} & \csc{0} & \csc{0.551}\\
CYP2D6 Veith              & AUPRC                   & \csc{0.566} & \csc{0} & \csc{0.641} & \csc{0} & \csc{0.487} & \csc{0.624} & \csc{0.704} & \csc{0.616} & \csc{0} & \csc{0.585}\\
CYP3A4 Veith              & AUPRC                   & \csc{0.523} & \csc{0} & \csc{0.649} & \csc{0} & \csc{0.713} & \csc{0.72} & \csc{0.818} & \csc{0.584} & \csc{0} & \csc{0.608}\\
CYP2C9 Substrate          & AUPRC                   & \csc{0} & \csc{0.523} & \csc{0} & \csc{0.558} & \csc{-0.377} & \csc{0} & \csc{-0.445} & \csc{-0.566} & \csc{0} & \csc{-0.586}\\
CYP2D6 Substrate          & AUPRC                   & \csc{0} & \csc{0} & \csc{0} & \csc{0} & \csc{0} & \csc{0} & \csc{0} & \csc{0} & \csc{0} & \csc{0}\\
CYP3A4 Substrate          & AUROC                   & \csc{0} & \csc{0} & \csc{0} & \csc{0} & \csc{0.409} & \csc{0.468} & \csc{0} & \csc{0} & \csc{0} & \csc{0}\\
Half Life Obach           & Spearman                & \csc{0} & \csc{0} & \csc{0} & \csc{0} & \csc{0.503} & \csc{0.671} & \csc{0.498} & \csc{0} & \csc{0} & \csc{0}\\
Clearance Hepatocyte      & Spearman                & \csc{0} & \csc{0.374} & \csc{0} & \csc{0} & \csc{0} & \csc{0} & \csc{0} & \csc{0} & \csc{0} & \csc{0}\\
Clearance Microsome       & Spearman                & \csc{0} & \csc{0.599} & \csc{0} & \csc{0.496} & \csc{0} & \csc{0} & \csc{0} & \csc{0} & \csc{0} & \csc{0}\\
LD50 Zhu                  & MAE                     & \csc{0.501} & \csc{0} & \csc{0.543} & \csc{0} & \csc{0.522} & \csc{0.586} & \csc{0.617} & \csc{0.339} & \csc{0} & \csc{0.342}\\
hERG                      & AUROC                   & \csc{0} & \csc{0} & \csc{0} & \csc{0} & \csc{0.57} & \csc{0.42} & \csc{0.453} & \csc{0} & \csc{0} & \csc{0}\\
AMES                      & AUROC                   & \csc{0.791} & \csc{0} & \csc{0.591} & \csc{0} & \csc{0.486} & \csc{0.629} & \csc{0.643} & \csc{0.604} & \csc{-0.628} & \csc{0.528}\\
DILI                      & AUROC                   & \csc{0} & \csc{0.376} & \csc{0.49} & \csc{0} & \csc{0.416} & \csc{0.567} & \csc{0.454} & \csc{0} & \csc{0} & \csc{0}\\
\midrule
& \multicolumn{1}{l}{Sum}                           & \csc{5.622} & \csc{4.883} & \csc{6.496} & \csc{2.269} & \csc{7.959} & \csc{9.712} & \csc{7.749} & \csc{2.499} & \csc{-2.232} & \csc{2.028}\\
\bottomrule
\end{NiceTabular}}
\vspace{-3mm}
\end{table*}

\subsection{Correlation analysis} \label{sec:correlation}

We also conduct a comprehensive correlation analysis to determine the impact of various pre-training datasets on the performance across 22 ADMET-group benchmarks from the TDC benchmark. We checkpoint on two critical points in the training process: the epoch with the lowest total validation loss and at the end of training (of which these two epochs can differ). This approach yielded 32 unique combinations of pre-training validation and downstream test metrics.

Spearman's rho correlation coefficients~\cite{sedgwick2014spearman} were calculated for each pair of pre-training and downstream metrics (see Tab.~\ref{tab:correlation_analysis}). The aim was to systematically identify which pre-training datasets have the most impact on downstream task performance, either positively or negatively. The p-value threshold was set at 0.1. Considering that some metrics indicate improvement when either increased or decreased, the correlation values were multiplied by the sign indicative of the direction in which improvement is registered for each metric. This adjustment provided a more nuanced understanding of whether enhancements in a pre-training metric are correlated with improvements in downstream performance.

We find an overall positive impact of our pre-training metrics on downstream performance thus demonstrating the effectiveness of \method pre-training step. Interestingly for graph-level tasks on \texttt{PCQM4M\_G25}, we find high pre-training results to be inversely correlated with strong performance on the ADMET group datasets. Specifically, a low MAE on \texttt{PCQM4M\_G25} results in negative effects across many downstream tasks. At the same time, we also find that the majority of pre-training metrics of \texttt{LargeMix} are highly informative of downstream performance, including the node-level tasks on PCQM4M. Further discussion is provided in~\Cref{sec:discussion_pcqm4m_g25} on possible explanations for these findings.

\section{Discussion}
Here, we discuss the differences in pre-training and fine-tuning performance for the different backbone GNNs of \method as well as a discussion on the results of our correlation analysis.

\subsection{Effect of model complexity}
For the fine-tuning results in~\Cref{sec:exp_fingerprints}, our analysis reveals that while the three GNN backbone, namely, GCN, GINE and MPNN++, all achieves similar pre-training performance, the GINE backbone shows a significant advantage when fine-tuning to downstream tasks. To give a potential explanation for this finding, recall that we adjust hidden dimensions of the different backbone GNNs to roughly align to 10M parameters. Here, the higher model complexity of MPNN++ leads to substantially smaller hidden dimension sizes than GINE. Our results thus indicate that the architectural complexity of the MPNN++ is less effective in downstream performance than a simple increase in hidden dimensions. If we match the hidden dimension size of GINE in MPNN, the model would reach the size of roughly 50M parameters. 

At the same time, while less complex and allowing for even larger hidden dimensions, GCN layers might be expressive enough for strong downstream performance. Specifically, the GCN omits the use of edge features and is shown to be less powerful than the 1-WL test~\citep{xu_how_2019} while GINE is as expressive as the 1-WL test. As such, we hypothesize that the GINE allows for a trade-off between a sufficient level of architectural complexity and more effective use of parameter budget in terms of larger hidden dimensions that translate into stronger downstream performance.

Finally, we want to highlight that our results also reveal the general robustness of \method pipeline to the choice of backbone GNNs, where all three variants were better than the current state-of-the-art foundation model MolE on the ADMET group in many tasks (as seen in Table~\ref{tab:models_on_tdc} and Table~\ref{tab:admet_original}) while having significantly fewer parameters and being employed in an efficient fingerprinting pipeline, as opposed to fine-tuning all weights.

\subsection{Detrimental effect of training on PCQM4M\_g25}\label{sec:discussion_pcqm4m_g25}

From the correlation analysis, results indicate that the PCQM4M\_G25 dataset negatively impacts downstream performance in the ADMET group benchmark. This dataset is unique in that it is the only graph-level quantum task.
As a result, we might conclude that atomic and sub-atomic properties of the molecules in PCQM4M\_G25 are only marginally relevant in the context of biological tasks, resulting in a detriment to performance on downstream tasks. 

However, this explanation is insufficient, so far that
the \texttt{PCQM4M\_N4} dataset, which is a node-level quantum task that covers the same molecules as \texttt{PCQM4M\_G25}, demonstrates positive downstream benefits. As a result, we hypothesize that the issue rather lies in the training dynamics. Concretely, models that perform well on \texttt{PCQM4M\_G25} might do so at the expense of overfitting, which is expected for multi-task learning on an unbalanced number of targets~\citep{beaini_towards_2023}. Indeed, as already described in \Cref{sec:exp_pre_training}, reducing the influence of PCQM4M\_G25 on the overall loss led to improved overall performance, supporting our hypothesis. 

Our findings highlight the importance of carefully designing the overall loss when pre-training on multi-task datasets such as \texttt{LargeMix}. Such a design should consider data imbalance to ensure positive impacts on downstream performance. This is particularly important for foundation models whose downstream applications are mostly unknown at the time of pre-training.

\section{Conclusion}

In this work, we propose a novel parameter-efficient foundation model for molecular learning called \method. \method is pre-trained on over 3300~biological and quantum tasks on graph- and node-level molecules and subsequently evaluated on the ADMET group of the TDC benchmark. \method outperforms the previous state-of-the-art foundation model on the ADMET group, MolE, with only 10M parameters, constituting only 10\% of MolE's 100M parameters. In addition, fine-tuning with MLPs on the fingerprints of pre-trained \method, allows for efficient fine-tuning.

Our empirical results showed that the molecular fingerprints extracted from \method are highly transferable to downstream tasks. \method achieves top 1 performance on 8 tasks and top 3 performance on 10 tasks on the ADMET group. In addition, our task correlation analysis highlighted the importance of carefully designing the overall loss for multi-task pre-training. With \method, we showcased the potential for parameter-efficient multi-task multi-level pre-training and fine-tuning on fingerprints.
One future direction is to design pre-training datasets that align with a wider variety of downstream tasks. We will open source \method and believe it will be an important tool for future research on molecular foundation models.

\section{Broader Impact}
Releasing our Model may not have immediate direct societal impacts. However, it is crucial to acknowledge the potential implications that arise when providing access to a foundation model for molecular graphs. One concerning possibility is the misuse of this technology for the development of chemical weapons, toxins, or unregulated drugs. To address these potential risks, we are committed to implementing robust mitigation strategies. Central to our approach is the active promotion of beneficial applications, particularly in the fields of material and drug discovery. By highlighting the positive utilization of this technology, we aim to channel its potential toward scientific advancements that contribute to societal well-being.

\newpage
\clearpage

\bibliography{ref}
\bibliographystyle{icml2024}
\newpage
\appendix
\onecolumn
\section{Appendix}

\subsection{MPNN architecture}\label{sec:mpnn_details}
\begin{figure}
    \centering
    \includegraphics[width=0.5\textwidth]{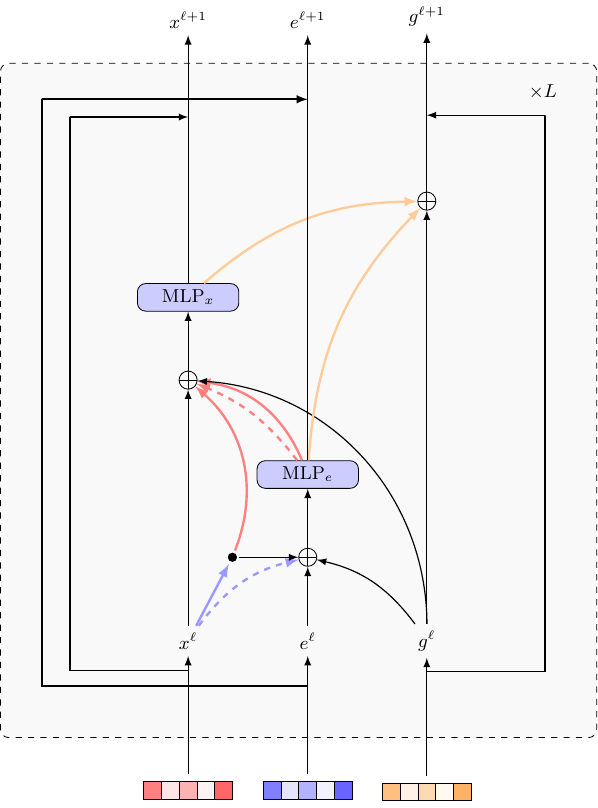}
    \caption{Example of the MPNN block architecture given in Eq.\ref{eq.mpnn}.  
    The edge update in Eq.\ref{eq:edge_update} gathers the nodes and edges before passing through the MLP first, then this output is used for the node update in Eq.\ref{eq:node_update}, gathering all connected node features and updated edge features. The global update in Eq.\ref{eq:global_update} connects all nodes and edges, before finally the skip connections in Eq.\ref{eq:skip}. 
     }

    \label{fig:mpnn}
\end{figure}

In what follows, we describe the MPNN architecture in \cite{masters_gps_2023} in detail.
Here, the embeddings are incrementally updated with each MPNN layer in the model  as:
 \begin{equation}
    x^{\ell + 1}, \, e^{\ell + 1}, \, g^{\ell + 1} = \texttt{MPNN}(x^\ell, \,e^\ell, \,g^\ell)
    \label{eq.mpnn}
\end{equation}
The edge embedding is updated by concatenation of the edge feature with the node features at each end of the bond, with the global features. This is processed with the edge MLP and then summed with the skip connection, shown in \ref{eq:edge_update}.  
 \begin{equation}
    \bar{e}^\ell_{uv} = \texttt{MLP}_{\textrm{edge}}\left(\,\left[x^\ell_u \,|\, x^\ell_v \,|\, e^\ell_{uv} \,|\, g^\ell\right]\,\right)
    \label{eq:edge_update}
\end{equation}
The node embedding, shown in eq.\ref{eq:node_update} concatenates the node features with the summed edge features of all edges connected (senders and receiver) and the global features before passing this vector through an MLP and finally adding the skip connection. 
 \begin{equation}
    \bar{x}^\ell_i = \texttt{MLP}_{\textrm{node}}\left(\left[ x^\ell_i \left | \sum_{(u,i) \in \mathcal{E}}\bar{e}^\ell_{ui} \right| \sum_{(i,v) \in \mathcal{E}}\bar{e}^\ell_{iv} \left| \sum_{(u,i) \in \mathcal{E}}x^\ell_{u} \right| g^\ell  \right]\right)
    \label{eq:node_update}
\end{equation}
 The global node is concatenated with the sum of all node and edge features in the graph (eq. \ref{eq:global_update}).
 \begin{equation}
    \bar{g}^\ell = \left[ g^\ell \left| \sum_{j \in \mathcal{V}} \bar{x}^\ell_j \right|  \sum_{(u, v) \in \mathcal{E}} \bar{e}^\ell_{uv}\right] 
    \label{eq:global_update}
\end{equation}
Where the final components are computed with skip-connections as:
\begin{equation}
    x^{\ell + 1}_i = \bar{x}^\ell_i + x^\ell_i; \quad e^{\ell + 1}_{uv} = \bar{e}^\ell_{uv} + e^\ell_{uv}; \quad g^{\ell + 1} = \bar{g}^\ell + g^\ell;
    \label{eq:skip}
\end{equation}
This is represented diagrammatically in Fig.~\ref{fig:mpnn}.

\subsection{Hyperparameter selection}
\label{sec:hyperparameter_selection}
We select hyperparameters for our fine-tuning as follows. We compute a hyperparameter sweep over the maximum number of epochs; the learning rate; the dropout rate and whether to use none, batch or layer normalization in the task head. Optionally, we sweep over the width and depth of the task head MLP. Each configuration is run on the same random seed. Following the instructions provided by TDC\footnote{Available at \url{https://tdcommons.ai/benchmark/overview/}}, we use the provided scaffold splits for our train/validation splits via the method $\texttt{get\_train\_valid\_split}$ and take the benchmark test split also provided by TDC.
Then, for each dataset, we select the hyperparameters resulting from the model with the smallest validation loss and subsequently re-run this model on $k$ random seeds. Here, we distinguish between two sweep configurations. In the first configuration, we only sweep over the learning rate $\in \{ 0.001, 0.0005, 0.0003, 0.0001, 5e^{-5}\}$ and set the number of epochs to $25$, dropout to $0.1$, hidden dimension to $1024$ and the number of layers to $3$. In the second configuration, we set the number of epochs to $25$ and sweep over whether or not to use a skip connection, the learning rate $\in \{ 0.0005, 0.0003, 0.0001\}$, the hidden dimension $\in \{ 512, 1024, 2048 \}$, the number of layers $\in \{3, 4\}$, dropout $\in \{0.0, 0.1\}$, the number of warmup epochs $\in \{ 0, 5\}$ and the learning rate schedule $\in \{ \text{constant}, \text{linear}, \text{cosine}\}$.

\newpage
\subsection{Experimentation with pooling methods} \label{sec:pooling_experiments}
We evaluated three different pooling strategies when going from the node level to graph level representation and summarized our findings in ~\ref{tab:pooling}.
\begin{table}[h]
\centering
\begin{tabular}{lcccc}
\hline
& \textbf{Mean rank} & \textbf{\>MoIE} & \textbf{TOP1} & \textbf{TOP3} \\
\hline
sum                & 3.9            & 16            & 5             & 10            \\
mean               & 3.8            & 16            & 6             & 11           \\
max                & 3.6            & 17            & 8             & 10           \\
\hline
\end{tabular}
\label{tab:pooling}
\end{table}

\subsection{Ensemble Strategy}
\label{sec:ensembling_strategy}
The strategy used for evaluating and ensembling the models is explained in the form of pseudo-code below.

\begin{algorithm*}[h]
   \caption{Ensembling Strategy}
   \label{alg:example}
\begin{algorithmic}
   \STATE {\bfseries Input:} hyperparameters $h_i$, repetitions $rep_i$, fold $fold_i$
   \FOR{each $h_i$ in $[hp_1, ..., hp_{18}]$}
   \FOR{$rep_i$ in $range(num_{reps})$}
   \STATE select $seed$
   \FOR{$fold_i$ in $range(num_{folds})$}
   \STATE train a model on $fold_i$
   \STATE save best model based on val loss
   \ENDFOR 
   \STATE build ensemble of $num_{folds}$ models
   \STATE evaluate on ensemble 
   \STATE save mean and std of val and test scores
   \ENDFOR
   \ENDFOR
\end{algorithmic}
\end{algorithm*}

\end{document}